\documentclass{article}
\usepackage{spconf,amsmath,amsfonts,graphicx}

\title{PCQA-GraphPoint: Efficient Deep-Based Graph Metric For Point Cloud Quality Assessment  }
\name{Author(s) Name(s)\thanks{Thanks to XYZ agency for funding.}}
\address{Author Affiliation(s)}

\name{Marouane Tliba$^{1}$, Aladine Chetouani$^{1}$, Giuseppe Valenzise$^{2}$ and Fréderic Dufaux$^{2}$}
\address{$^{1}$Laboratoire PRISME, Université d'Orléans, Orléans, France\\
$^{2}$Université Paris-Saclay, CNRS, CentraleSupélec, Laboratoire des signaux et systèmes
\\}

\begin{document}
%
\maketitle
\begin{abstract}
Following the advent of immersive technologies and the increasing interest in representing interactive geometrical format, 3D Point Clouds (PC) have emerged as a promising solution and effective means to display 3D visual information. In addition to other challenges in immersive applications, objective and subjective quality assessments of compressed 3D content remain open problems and an area of research interest. Yet most of the efforts in the research area ignore the local geometrical structures between points representation.  In this paper, we overcome this limitation by introducing a novel and efficient objective metric for Point Clouds Quality Assessment, by learning local intrinsic dependencies using Graph Neural Network (GNN). To evaluate the performance of our method, two well-known datasets have been used. The results demonstrate the effectiveness and reliability of our solution compared to state-of-the-art metrics.
\end{abstract}
\section{Introduction}
\label{sec:intro}

Recently, with the increasing demand for immersive applications and the widespread adoption of 3D acquisition devices, 3D Point Cloud (PC) data becomes more and more popular due to its capability to display the real world in modern immersive communications systems. Point clouds are becoming more widely used in a variety of fields, such as robotics, 3D gaming and telepresence. PCs consist of unordered geometrical coordinates $(x,y,z)$ as well as potentially associated attributes such as color, curvatures, reflectance and normal vectors. Consequently, PCs can accurately represent a 3D model or an entire scene by incorporating thousands or even millions of points, requiring efficient compression schemes. To optimize the coding of point clouds and obtain the best visual quality for a given bitrate, accurate perceptual quality metrics are needed. Hence, point cloud quality assessment (PCQA) has received a great deal of attention from the research community in recent years. 


A number of point cloud quality metrics have been developed in the past few years. They can be classified into three main groups: Point-based, Feature-based and Projection-based metrics. Point-based metrics such as Point-to-Point (Po2Po) \cite{Po2Po}, Point-to-Plane (Po2Pl) \cite{Po2Pl}, Plane-to-Plane (Pl2Pl) \cite{Pl2Pl} and Point-to-Mesh (Po2Mesh) \cite{Po2PM}, predict the quality through point-wise geometric and/or features distance between the reference PC and its distorted version. Po2Po measures the relative distance between point pairs to estimate the final quality, Po2Pl extends Po2Po by projecting the error vector along the local normal, and Pl2Pl quantify the quality through measuring the angular similarity between surfaces associated to the points from the reference and degraded contents. Po2Mesh creates a polygonal mesh from the reference sample and then compute the distance between each distorted point and the corresponding surface. Currently, MPEG is adopting Po2Po MSE and Po2Pl MSE with the associated PSNR as the standard point cloud geometry quality metrics.
Feature-based PC quality metrics extract the geometry with the associated attributes from point-wise level in a global or local way. Among those metrics, we can cite PC-MSDM \cite{PC-MSDM} that extends the 2D SSIM metric \cite{SSIM} to PC  by considering local curvature statistics, the Geotex \cite{Geotex} metric that exploits the Local Binary Pattern (LBP) \cite{LBPSurvey} descriptors, and PCQM \cite{PCQM} that combines the geometry and color features.
In projection-based PC Quality Metrics, the 3D points or their associated features are projected into 2D regular grids, and later, 2D methods are applied on these views \cite{Salima}, or 2D grid of features \cite{AladineNoRef} \cite{AldineWithRef}.
\begin{figure*}[t]
  \centering
  \centerline{\includegraphics[scale=0.725 ]{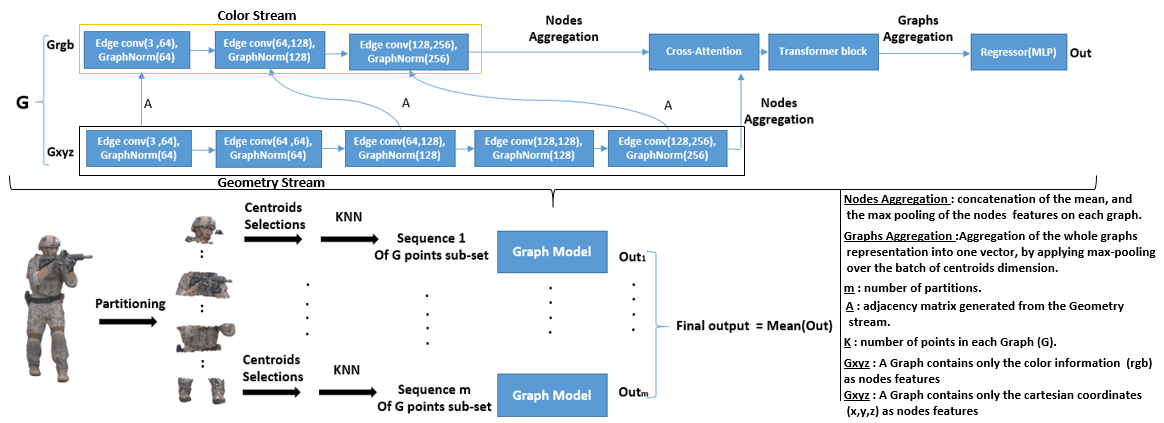}}

\caption{ General pipeline of our proposed method }
\label{fig:gen_arch}
\end{figure*}
The above point-based metrics or projection based metrics require a lot of pre-processing time, as well as the reference point clouds. Recently, new efficient deep based metrics such as
 PointNet-SSNR 
 \cite{ICIPPOINTNET} and PointNet-DCCFR\cite{tliba2022point} have been released. They exploit the intrinsic features of point cloud data, by operating on the native point cloud directly without applying any projection or transformation. 
 A main limitation of these methods is that they mainly use point-based convolutions to extract features from points, disregarding the local relationships among points.


To overcome this limitation, we propose in this paper a novel efficient end-to-end deep graph-based metric for no reference point cloud quality assessment that considers the structure of the point cloud. Unlike previous methods, we take into account geometrical local connectivity between points, as well as hidden deep representation dependencies. 
Our method is designed to capture efficiently geometrical local point cloud structures through a dynamic graph construction at each network layer, and improves the aggregation of distant point dependencies in the latent space using cross/self-attention mechanisms. 
In this paper we demonstrate the effectiveness of our approach in estimating the visual quality of point clouds corrupted by compression artifacts, which is one of the main distortion scenarios in practice. 
The main contributions of this paper are summarized as follows:

\begin{enumerate}
   \item We propose a novel efficient end-to-end deep graph based metric for PCQA, that operates on the whole point cloud  without any projection or other transformation.
   \item Our method is designed to capture efficiently geometrical local point cloud structures through the dynamic graph construction at each layer, and draws better aggregation of distant dependencies on the latent space using cross/self-attention mechanism. 

\end{enumerate}

\section{Proposed Method}
\label{sec:method}
The main contribution of this work is to provide an efficient and robust blind metric for estimating the quality of 3D compressed PCs. Specifically, we propose a graph-based metric considering local connectivity information between adjacent point representations.  The overall pipeline of our method consists of several steps: pre-processing, features extraction and aggregation using dynamic graph convolutional neural network (DGCNN), attention mechanisms and finally quality estimation. The general pipeline is depicted in Fig. \ref{fig:gen_arch}.

\subsection{Pre-processing}

The pre-processing stage consists in dividing the PC into vertical slices of points. This partitioning has two goals: on one hand, to enable parallel processing of points, and on the other hand, to meet memory requirements for GPU processing. The number of partitions varies based on the size of the original PC. In our tests, we use between 8 and 24 partitions. Each new point set (partition) is then divided into subsets in order to form a local graph on each of them. To do so, we select a random number of (distant) centroids, and apply the $k$-nearest neighbor clustering method to form a patch around each centroid. 
Each point subset is then fed into our \textit{graph model} for graph construction and features extraction. 

\subsection{Proposed Graph Model}

Our proposed model is mainly inspired from Pointnet++ \cite{PointNetDL} and DGCNN \cite{DGCNN}. In particular, we leverage the information from local structures by building a graph based on neighboring points, and apply convolution using the graph edge information. Also, similar to DGCNN \cite{DGCNN}, we update dynamically the graphs after each layer of the network, to better capture the dependencies at different feature levels. However, unlike DGCNN that is designed to extract semantic features for high-level computer vision tasks, our model employs graphs only to extract local features, and uses instead self/cross attention to process semantic information. More specifically, we use two-stream graph neural networks, one for the geometrical information, and one for color information. At a later stage, we aggregate the two streams using a cross-attention mechanism. 

In more details, at each layer of the network, we construct a graph using $k$-nearest neighbors ($k=6$) of a point. In the first layer, the graph is built using the geometric distances among points. At subsequent layers, the graph is constructed using distances between \textit{embeddings} of points at that layer, thus effectively capturing local structures in the feature space. This results in a new adjacency matrix computed at each layer. We use the adjacency matrices of the geometry stream also for the color stream, in order to align the color information propagation with local geometrical structures. 

\subsubsection{Two-stream graph network}

%
We consider geometry and color information as features, corresponding to two different processing streams. Depending on which stream we consider, input points $\mathbf{x}_i$ bring different information. For the geometry stream, $\mathbf{x}_i = (x_i, y_i, z_i)$ contains the 3-dimensional coordinates of the points, while for the color stream, points represent the RGB attribute information. Notice that it is possible to include additional features in other streams.

We construct two directed graphs  by finding the $k$-nearest neighbors of each point ${G_{xyz}}$ = ($\mathcal{V}_{xyz}, \mathcal{E} )$, and ${G_{rgb}}$ = ($\mathcal{V}_{rgb}, \mathcal{E} )$, representing  point cloud local structures along side the geometric and color features, where $\mathcal{V} = \{1, \hdots, k\}$ and $\mathcal{E} \subseteq \mathcal{V} \times \mathcal{V}$ are the \textit{vertices} and \textit{edges}, respectively. Each graph is also completed with self-loop. 

\begin{figure}[t]
  \centering
  \centerline{\includegraphics[width=\columnwidth ]{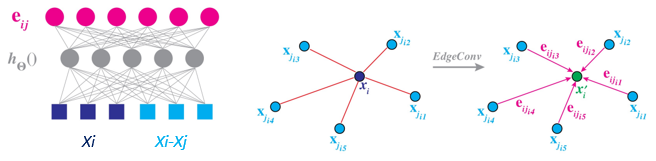}}
\caption{\textbf{Left}: Computing an edge feature  $\boldsymbol{e}_{ij}$, \textbf{Right}: The EdgeConv operation \cite{DGCNN}.}
\label{fig:edge}
\end{figure}

The \textit{edge features} at each network layer of the two streams are defined as an asymmetric function  $\boldsymbol{e}_{ij} = h_{\boldsymbol{\Theta}}(\mathbf{x}_i,\mathbf{x}_i - \mathbf{x}_j)$, where ${\mathbf{x}_j}$ is the
central vertex, $\{ \mathbf{x}_j : (i,j) \in \mathcal{E}\}$ is the neighbor vertex, and $h_{\boldsymbol{\Theta}} : \mathbb{R}^3\times \mathbb{R}^3 \rightarrow \mathbb{R}^{F}$ represents a nonlinear function with learnable parameters $\boldsymbol{\Theta}$. The adopted convolution operation, so-called \textit{Edge Conv}, as depicted in Fig.\ref{fig:edge}, is defined  by applying a $\max$ channel-wise aggregation of the features associated with all the edges resulted from each vertex. Therefore, the output  at the $i$-th vertex of each stream is given as %
\begin{equation}
\label{eq:full}
\mathbf{x}'_i  =\mathop{{\max}}_{j: (i,j) \in \mathcal{E}} h_{\boldsymbol{\Theta}}(\mathbf{x}_i, \mathbf{x}_i - \mathbf{x}_j).
\end{equation}

In order to have faster convergence, after each \textit{Edge Conv} layer we applied \textit{GraphNorm} \cite{graphnorm} operation, an adapted layer of InstanceNorm \cite{Ulyanov2016InstanceNT} for graph normalization with a learnable parameter $\alpha$  that controls how much of the channel-wise average to preserve in the shift operation:
\begin{equation}
\mathbf{x}^{\prime} = \frac{\mathbf{x}^{\prime} - \alpha \cdot
\mathbb{E}[\mathbf{x}^{\prime}]}
{\sqrt{\textrm{Var}[\mathbf{x}^{\prime} - \alpha \cdot \mathbb{E}[\mathbf{x}^{\prime}]]
+ \epsilon}} \cdot \gamma + \beta
\end{equation}
where ${\gamma}$ and, ${\beta }$ are the affine learnable parameters similar to other normalization methods.

After multiple consecutive (\textit{Edge Conv}, \textit{GraphNorm}) layers on each stream, a \textbf{Nodes Aggregation} is applied on each graph  independently. Here a Max and Mean pooling operation  are applied on node channel dimension. Afterwards, the result of the two pooling is concatenated to one  vector, so each graph $G$, transformed into sequence of ${2F}$-dimension vectors, denoted as  ${Hidden}$.

\subsubsection{Two Streams Fusion}

The representations produced by the two streams,  ${Hidden}_{rgb}$ and ${Hidden}_{xyz}$, are finally fused  using a cross-attention layer \cite{cross-vit}. More precisely, the fusion is carried out by measuring the similarity between the queries $\mathbf{q}$ (a transformed vector of ${Hidden}_{rgb}$) and the keys $\mathbf{k}$ (a transformed vector of   ${Hidden}_{xyz}$) as well as using it to adjust the values vector $\mathbf{v}$ (a transformed vector of   ${Hidden}_{xyz}$). Mathematically, the cross-attention can be expressed as follows:
\begin{gather} 
\mathbf{q} = {Hidden}_{rgb}W_q ,\mathbf{k} = {Hidden}_{xyz}W_k,\mathbf{v} = {Hidden}_{xyz}W_v \\
\mathbf{A} = \frac{softmax(\mathbf{q}\mathbf{k}^{T})}{\sqrt\frac{{C}}{h}}\\ {CrossAttention}({Hidden}_{xyz},{Hidden}_{rgb}) = \mathbf{A}\mathbf{v}
\end{gather}
where ${W_q}, {W_k}, {W_v} \in {R}^{C×(C/h)} $
are learnable parameters, ${ C }$ and ${ h} $ are the embedding dimension and number of heads. We refer the reader to \cite{transformer,cross-vit} for more details about the computation of cross-attention.

\subsubsection{Feature Aggregation and Quality Estimation}

To better aggregate the fused representation and draw the relationship between distant embedded features on the output of the fusion stage, we employed also a self-attention layer~\cite{transformer}. Unlike cross attention~\cite{cross-vit}, here the set of $\mathbf{k}$, $\mathbf{q}$, and $\mathbf{v}$ are belonging to the same distribution of features~\cite{satsal}. Following to that, similar to transformer block~\cite{transformer}, we add skip connections with layer normalization, followed by a non-lineaar function. 

Finally, we use a \textbf{Graphs Aggregation} layer, which applies a max poling operation over the vector output by the transformer block, in order to have one vector that represents each PC partition. The later is passed to a very shallow multi-layer perceptron  (i.e. regressor) in order to estimate the quality score of each PC partition. The final quality score of the whole PCs is obtained by applying the mean on the sequence of partitions scores.

\subsubsection{Training Protocol and Implementation details}

We train our model in an end-to-end manner, using the mean square error (MSE) as a loss function. The network is trained to create a mapping function between the input PCs and the mean opinion score (MOS) quality. The loss function is defined as follows: 
\begin{equation}
{
\mathcal{L}= \mathrm{MSE}\left(\mathrm{mean}\left(\sum_i{Out_{i}}\right), \mathcal{Y}\right).
}
\end{equation}
where $Out_{i}$ refers to each partition predicted score, and $\mathcal{Y}$ refers to the MOS.

We implemented our model in PyTorch. The model was trained using Adam optimizer \cite{Adam} with initial learning rate of 0.0001 and a batch size set to one.


\begin{table}[t]
\small
\label{tab:tab1}
\caption{\label{tab:icip_results} Results obtained on ICIP20 dataset using 6-fold cross validation  
}
\begin{center}
\begin{tabular}{ c c c  }

\hline
\textbf{Model} & \textbf{PLCC  $\uparrow$ } & \textbf{SROCC $\uparrow$ }   \\ 
\hline
 po2point MSE & \textbf{ 0.946} & 0.934     \\ 
 \hline
 po2plane MSE & \textbf{0.959} & 0.951         \\
 \hline
 PSNR po2point MSE & 0.868 & 0.855         \\
 \hline
PSNR po2point HAU & 0.548 & 0.456 \\
\hline 
PSNR po2plane HAU  &0.580 & 0.547\\
\hline 
color Y MSE  &0.876  &0.892 \\
\hline
color Cb MSE &0.683& 0.694\\
\hline
color Cr MSE &0.594 &0.616 \\
\hline
color Y PSNR& 0.887& 0.892 \\
\hline
color Cb PSNR &0.693& 0.694\\
\hline 
color Cr PSNR &0.626& 0.617 \\

\hline
pl2plane AVG &0.922 &0.910 \\
\hline
pl2plane MSE &0.925& 0.912 \\
 \hline
 PCQM & 0.796 & 0.832       \\ 
\hline
 PointNet-SSNR &  {0.908} &  {0.955} \\
 \hline
 PointNet-DCCFR &  \textbf{ 0.947} &  \textbf{0.973} \\
 \hline
Our Color stream &  { 0.836}  & {0.925} \\
 \hline
 Our Geometry stream &  \textbf{ 0.925}  & {0.939} \\
\hline
Our &  \textbf{ 0.946}  & \textbf{0.973} \\
 \hline
\end{tabular}
\end{center}
\vspace{-9mm}
\end{table}



\section{Experiments}

To validate the effectiveness of our model, we evaluate it on two publicly available benchmarks with subjective scores, which adopt different emergent  compression schemes with multiple encoding levels: ICIP20 \cite{ICIP20} and PointXR \cite{pointxr}.
\textbf{ICIP20} is composed of 6 reference PCs from which 90 degraded versions were derived through three types of compression: V-PCC, G-PCC with triangle soup coding and G-PCC with octree coding \cite{GPCC,Octree,TreeSoup}. Each reference PC was compressed using five different levels. \textbf{PointXR} is composed of 5 PCs from which 45 degraded versions were derived through G-PCC with octree coding for geometry compression and, Lifting and RAHT for color compression.

To train and test our model on ICIP20, we split the dataset according to the reference point cloud by applying a 6-fold cross validation protocol where 6 corresponds to the number of reference point cloud samples, i.e., at each iteration $5$ reference point cloud samples and their compressed versions are used for training, and one reference point cloud sample and its compressed versions are used for testing. Then, in order to evaluate the generalization capacity of our method to predict the quality on unknown PCs, a cross-dataset evaluation is employed by training our model on ICIP20 and testing it on PointXR datasets. Pearson Correlation Coefficient (PCC) and Spearman Rank-Order Correlation Coefficient (SROCC) are considered as criteria for measuring the quality prediction ability of our method. Notice that the correlations are computed for each fold and the mean over all folds is reported. The results obtained are shown in Tables 1 and 2. We compared the performance of our method to state-of-the-art methods by applying strictly the same protocol (i.e. splitting and reporting the mean correlations). 

In Table~\ref{tab:icip_results}, we show the performance reached by our method on ICIP20 dataset for the two streams independently, denoted here as Color stream and Geometry stream, as well as after considering the two streams (i.e. fusion). The results are also compared to state-of-the art methods. As we can see, the two streams are well correlated with the subjective ground truth. The geometry stream predicts the quality much better than the color one, especially according to PLCC. After combining the two streams, the performance improves with a clear gap for both PLCC and SROCC. Compared to the existing methods, the proposed one outperforms most of them with the best SROCC and the second best PLCC. It is worth noting that our method surpasses most of the full-reference methods. Notice also that all the listed methods in the table are full-reference except PointNet-SSNR\cite{PointNetDL}, which is no-reference. 

\begin{table}[t]
\small
\label{tab:tab2}
\caption{\label{tab:PXR}Results obtained by training the model on ICIP20 and testing it on PointXR}
\begin{center}
\begin{tabular}{ c c c  }
\hline
\textbf{Model} & \textbf{PLCC  $\uparrow$ } & \textbf{SROCC $\uparrow$ }   \\ 
\hline
 po2pointMSE &   0.887 & 0.978  \\ 
 \hline
 po2planeMSE & 0.855 & 0.942  \\
 \hline
 PSNRpo2pointMSE & \textbf{0.983} & 0.978  \\
 \hline
 PSNRpo2planeMSE & 0.972 & 0.950  \\
 \hline
 PointNet-DCCFR  & 0.981 &  {0.964}    \\
 \hline
Our  & 0.967&  \textbf { 0.988 }    \\
 \hline
\end{tabular}
\end{center}
\vspace{-7mm}
\end{table}

Table~2 shows the performance obtained for the cross-dataset evaluation. As can be seen, high correlations are achieved by our method, outperforming some of the compared ones, showing the generalization capability of our metric to predict the quality of unknown PCs. These results show that the proposed method has consistent performance across different validation sets.


\section{Conclusion}
In this paper, we proposed a novel learning-based no-reference quality metric for point clouds. 
Unlike previous learning-based approaches, we employ graph convolutions to extract local features at different network layers. Specifically, we consider a two-stream architecture to process in parallel geometry and color distortion, and we use cross- and self-attention to combine features leveraging long-term dependencies in the point cloud. Our proposed method achieves state-of-the-art correlations on two benchmark datasets. In the future, we believe that this approach should be further validated as larger annotated datasets for point clouds quality become available.

\label{sec:conclusion}

\bibliographystyle{IEEEbib}
\bibliography{strings,refs}

\end{document}